\documentclass[letterpaper]{article}
\usepackage{aaai}
\usepackage{times}
\usepackage{helvet}
\usepackage{courier}
\usepackage{graphicx}
\usepackage{multirow}
\usepackage{tabularx}
\usepackage{amsmath}
\usepackage{subcaption}
\usepackage{natbib}
\usepackage{hyperref}
\usepackage{shortcuts}

\usepackage{graphics}
\DeclareMathOperator{\argmax}{arg\,max}

\begin{document}
\title{Human-like Time Series Summaries via Trend Utility Estimation}
\author{Pegah Jandaghi \\
University of Southern California \\
Los Angeles, CA 90007 \\
jandaghi@usc.edu
\And
Jay Pujara \\
Information Sciences Institute \\
Marina del Rey, CA 90292 \\
jpujara@usc.edu
}
\maketitle

\begin{abstract}
\begin{quote}
In many scenarios, humans prefer a text-based representation of quantitative data over numerical, tabular, or graphical representations. The attractiveness of textual summaries for complex data has inspired research on data-to-text systems. While there are several data-to-text tools for time series, few of them try to mimic how humans summarize for time series. In this paper, we propose a model to create human-like text descriptions for time series. Our system finds patterns in time series data and ranks these patterns based on empirical observations of human behavior using utility estimation. Our proposed utility estimation model is a Bayesian network capturing interdependencies between different patterns. We describe the learning steps for this network and introduce baselines along with their performance for each step. The output of our system is a natural language description of time series that attempts to match a human's summary of the same data. 
\end{quote}
\end{abstract}

\section{Introduction}
There is a vast amount of data, and understanding this data presents a cognitive barrier for people. Studies show that in many scenarios, people prefer a text description of data over numerical, tabular, or graphical representations of it. As an example, medical staff made better treatment decisions when presented with a text description of patient status compared to graphs~\citep{Law2005ACO}. In this paper, we present an approach to generate textual summaries using a probabilistic model that represents the complex patterns of human summarization.

Diverse data-to-text systems have been proposed for generating summaries~\citep{datatotext}.
Unfortunately, most efforts to automatically generate text descriptions of data fail to consider which aspects of the data are most important to a human end user.
In this paper, we introduce our summary generation system for numerical time series. The goal of this system is to learn how humans describe time series data and create a descriptive natural language text similar to the human summary. 

Human summaries can capture many different features from data, including relationships to background knowledge and comparisons to other, unseen, data.  In this paper, we focus on simulating parts of the text that are directly or indirectly describing a salient pattern in the data. To do so, we try to learn common numerical patterns used by people, the various textual statements that are used in describing them, and how they are aligned with each other.
This process leads to finding interpretable patterns in data which we call \emph{trends} and their textual descriptions which we use as \emph{templates}. For example, in the sentence ``TSLA stock has plummeted 15 percent in the past three months'' the verb ``plummeted'' signals a sharp decreasing trend in the time series.

We detect these trends from numerical time series data, propose a utility estimation model to detect a subset of commonly used trends that are present in time series and learn when and how these trends are used by humans. 
The core idea in our model are a set of \emph{policies}, which represent latent variables parameterized by data features that dictate when a trend is included in a summary. To model the complex interactions of summarization policies, we use a Bayesian network.
The output of the system consists of a set of templates, each of which is associated with a high utility pattern in the data.

\section{Problem Statement} 
\begin{figure*}
\centering
\begin{subfigure}{.95\columnwidth}
  \centering
  \includegraphics[width=.9\linewidth]{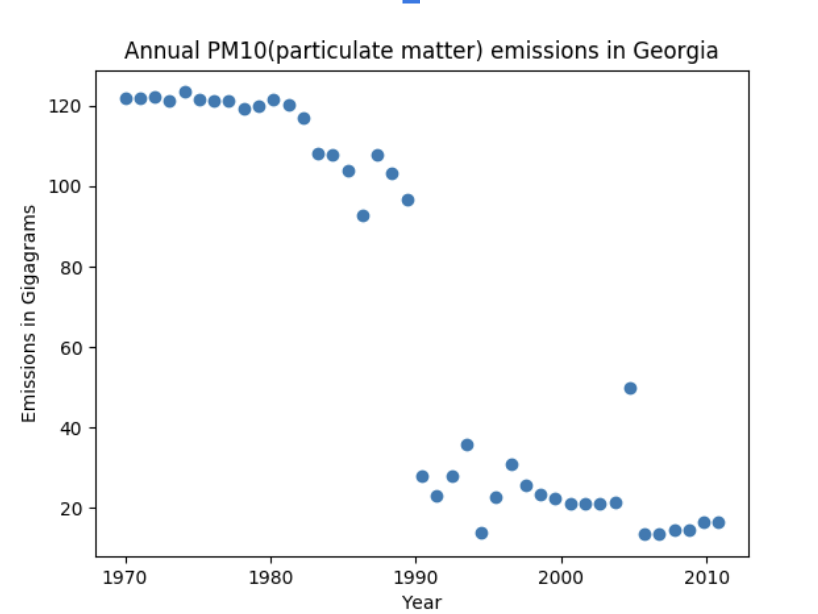}  
  \caption{A set of trends in this time series includes: 1) A gap of size 60 between 1989 and 1990 2) A anomalous point in 2005 compared to the value of time series in 2005 to 2010}
  \label{fig:sub-first}
\end{subfigure}
\begin{subfigure}{.95\columnwidth}
  \centering
  \includegraphics[width=.9\linewidth]{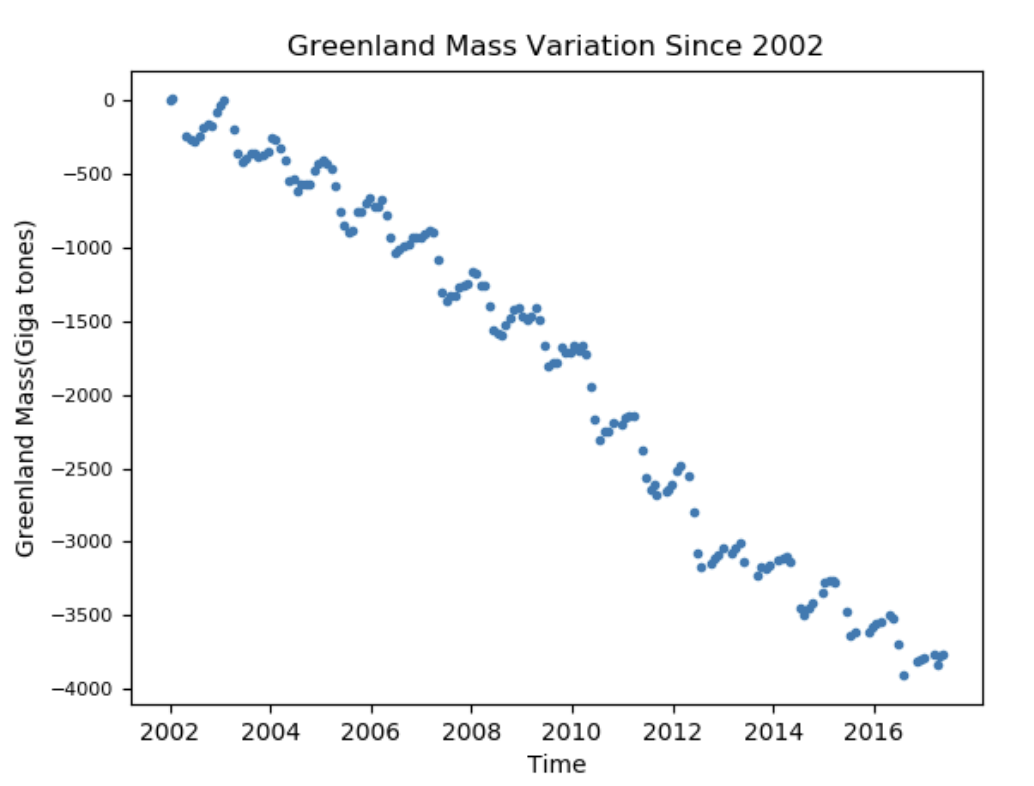}  
  \caption{A set of trends in this time series includes: 1) A cycle pattern from 2002 to 2018 with yearly period 2) a linear decreasing pattern from 2002 to 2018}
  \label{fig:sub-second}
\end{subfigure}
\caption{Prevalent trends in time series examples.}
\label{fig:trends}
\end{figure*}

In this section, we formalize our summary generation model for numerical time series data. Our model is based on identifying prominent trends in data and creating textual descriptions for them. A \emph{trend} is a pattern in data which is interpreted by a human and can be qualitatively described in text. As an example, \figref{fig:sub-second} shows a dataset of Greenland mass variation, with a cyclic pattern corresponding to a trend which has been described in the sentence "These oscillations are waves of mass variation which occurred on an annual or biennial basis". 
A trend can be a value of a point or set of points in the data (such as a maximum), a relationship between points or sets of points (such as an increasing trend), or an aggregate measure on a subset of data points (such as a mean value). Although there are many possible patterns in the data, we observed that certain categories of patterns appeared  more frequently in human-generated summaries of data, which are our main focus in this paper. These trend categories are linear, statistical properties, discontinuous transitions, cycles, and anomalous points. Figure \ref{fig:trends} contains examples of these trends with their descriptions.\\

We now provide a formal definition of time series trends, their features, and the utility of a trend.
Suppose $ts$ is an arbitrary time series. We represent each trend $tr$ observed in $ts$ with an m-element feature vector $v_{tr} \in R^m$. A feature vector for each trend contains parameters needed to describe the trend in a textual summary and approximately reconstruct the underlying data pattern. For example, the feature vector for the  a linear trend contains the slope, intercept, the spanning interval, the number of data points, etc. These features can be general or specific to a trend type (e.g., the slope feature is defined for linear trends whereas the spanning interval is defined for all trends).

Let $X$ be the set of all possible trends and $V$ be the associated features for these trends, s.t. $\forall tr \in X, v_{tr} \in V$. We introduce a utility model which captures the preference of trends.
The general utility model tries to find a utility function $\forall T \subset X, V_{T} \in V, U:\mathcal{P} (T,V_T) \rightarrow [0,1]$ which maps a set of trends and their associated features to a utility value in the $[0,1]$ interval for the summary consisting of that set of trends. The utility function imposes a weak ranking over all possible subsets of trends, creating a preference function. Note that in practical applications, utility functions may be personalized to specific populations or subject domains.

The general form of this utility model leads to a potentially intractable problem. Datasets contain many trends, and the utility function must be computed over the powerset of all such trends. 
In this paper, we focus on a simpler problem that provides utilities for individual trends. We introduce simplified version of utility model $U':(X,V) \rightarrow [0,1]$ that assigns utilities to individual trends.

\section{Proposed Model}

In this section, we formally describe our utility model, our approach of introducing latent summarization policies to model human behavior, and define the associated tasks for learning our model.

\subsection{Utility Model}
In this section, we describe our utility model. Let $ts$ be an arbitrary time series and $tr$, $tr'$ be trends in $ts$ with feature vectors $V_{tr}, V_{tr'} \in R^m$. Let $\overline{V_{tr}}$ be feature vectors of all trends in $ts$ except $tr$. Let $Y_{tr}$ be a binary random variable indicating whether $tr$ appears in the text or not.
We identify the utility of a trend $tr$ as the probability that we observe $tr$ in the text that is $P(Y_{tr} = 1)$.
Our goal is to learn probability distribution of $Y_{tr}$ based on the observed data which is $P(Y_{tr}|V_{tr}, \overline{V_{tr}})$. We propose a graphical model to estimate this distribution. In our model, observed variables are $V_{tr}$, $\overline{V_{tr}}$ and during training we are given $Y_{tr}$. The hidden variables in this model are latent policies, described in the next section. 

\subsection{Policies} \label{policies}
A policy is a binary random variable whose distribution depends on the trend feature vector. The value of the policy indicates whether a trend is selected for the summary based on its feature vector and other trends in that series. For example let $\pi_x$ be the policy that prefers more recent trends. In other words $\pi_x$ considers the attribute of trends which refers to their spanning interval. The value of $\pi_x$ is more likely to be 1 for the most recent trend and 0 for initial trends.
In this paper, we define a compositional model of utility that defines the utility of complex preferences as a combination of simpler models, which we refer to as policies. The primary difference between policies and the utility model is that policies are explanatory latent variables for a trend, while the utility function estimates the empirical probability of a trend aggregated over a dataset.

We divide policies into leaf policies and complex policies. A leaf policy is an atomic policy that can not be decomposed as a set of policies combined with binary operations, considering only one aspect of the trend or trends and its value for each trend is independent of other policies. 
For example the policy $\pi_x$(introduced in the previous paragraph) that prefers more recent trends is a leaf policy. Complex policies can be created by combining leaf policies using binary logical operations i.e conjunction, disjunction, exclusive or etc. For example, let $\pi_y$ be the policy that is 1 when the the linear trend is increasing. The policy that prefers the most recent increase in a time series can be viewed as $\pi_x \wedge \pi_y$. \\
The criteria used in leaf policies may vary from simple to complex. The criteria may use limited features of trends or might consider dependencies among different features in different trends. We gathered a set of criteria that humans used in their preference models and classified leaf policies into following categories based on them.
\begin{itemize}
	\item Single Feature: In this policy category, the value of the policy depends on the value of a single feature. We assume that in this case, the value of policy is derived from a simple function of that feature, For example, a threshold function measures when a feature value exceeds a threshold can be used to define a policy that selects linear increasing trends by setting a threshold of 0 on the slope feature. 
	\item Multiple Features: In this policy category, the value of a policy depends on multiple features of a single trend. For example, this policy can be used to define cases when a linear trend has a slope greater than the intercept value. 
	\item Single Feature in Multiple Trends: In this policy category, the policy value for a trend depends on a single feature of that trend as well as the same feature in other trends in the time series. This policy type can be used to compare trends. For example, the policy that prefers the most recent trend is in this category since it requires comparing the "interval" feature of all trends in the same time series. 
	\item Multiple Features in Multiple Trends: In this policy category, the policy value for a trend depends on multiple features of the trend as well as other trends. For example, a policy that prefers jump points that do not exceed 50\% of the maximum value of a time series fall into this category. 
	\item Feature Independent: In this type of policy, the value of policy is not determined by feature vector of trends. In this case, a series of hidden factors affect the utility of trends, e.g., a hidden factor might be the context of the time series. Note that we do not consider this leaf policy category in our model.
\end{itemize}
The leaf policies can be combined using different logical structures to create various complex policies. Complex policies may have different and conflicting values for trends. For example a complex policy might have a high value for a specific trend whereas another policy might have low value for the same trend.
Although there exist many complex policies, when and how these policies are activated depends on the specific summarization context, and some policies are not considered in assessing the utility of some trends. 
Therefore, the utility of each trend is dependent on a specific subset of these policies and each of them might have different degree of importance. For example, suppose $tr$ is a linear increasing trend in stock indicator $X$ spanning from 2009 to 2013. A policy identifying long-running trends may be triggered by this trend, while a policy that identifies recent data may ignore this trend.
A utility model learns that the second policy is a more reliable indicator of human behavior than the first policy may then omit this trend from a summary.
\\
Our goal is to find the utility function that estimates the utility of the trends by assigning high utility values to the trends that human prefer. The problem of finding the utility function can be formulated as finding the leaf policies, finding complex policies which requires determining the structure of dependency between complex policies and leaf policies and finding the joint distribution of complex policies for different trends. 
\begin{figure}[t]
\includegraphics[width=\columnwidth]{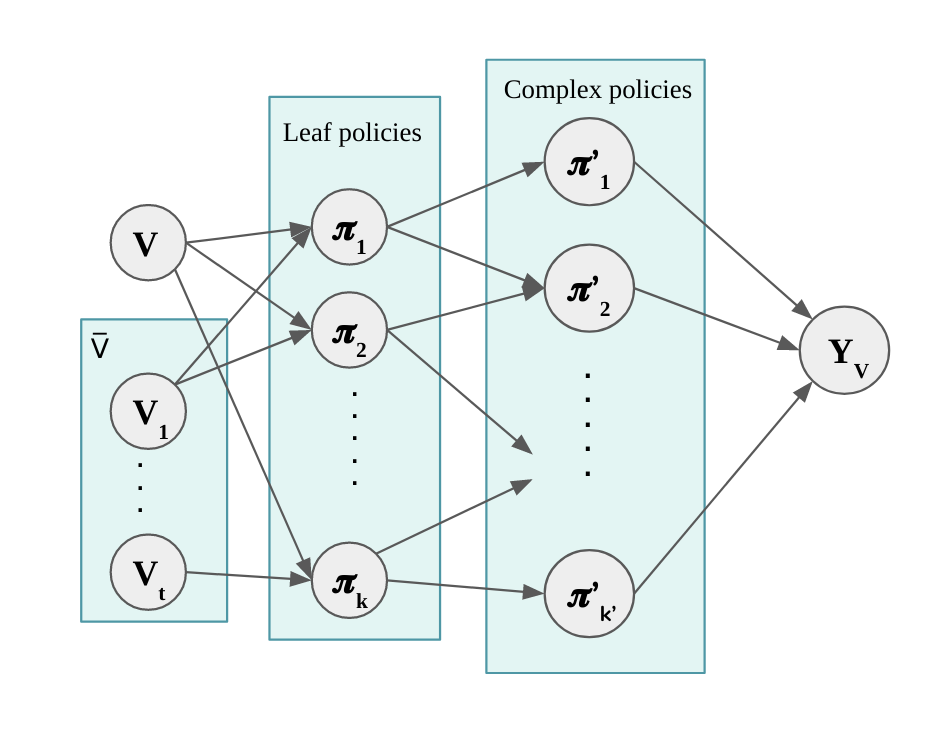}
\centering
\caption{The architecture of our utility model}
\label{arch}
\end{figure}

\subsection{Using Policies for Utility Estimation}
The architecture of our utility model is a Bayesian network shown in figure \ref{arch}. In this paper, the simplified task is restricted to predicting the utility of a single trend. Simple policies are defined with respect to feature vectors for the target trend as well as other trends. These policies are combined using logical formulae to create complex policies. During utility estimation, the output variable of whether a trend is included in the summary is defined using a probability distribution over the complex policies.

We define leaf policies to be binary random variables.
Let $\pi_1, .., \pi_k$ be the leaf-policies in the model.
Complex policies are created using different structures and arithmetic logic on leaf-policies.
For example let $\pi_1$ and $\pi_2$ be leaf-policies they can be combined using xor and create a new policy $\pi'$. Let $S_1, S_2, ..., S_{k'}$ represent the structures that are used in the model, therefore $k'$ complex policies are present in the model which we denote by $\pi_1', .., \pi_{k'}'$. Complex policies are also binary random variables and are dependent on leaf policies. The value of each complex policy is independent of other policies and the value of $Y_{tr}$ is dependent on all complex policies. Therefore we can compute the final utility of trend $tr$ in our model as:
\\
$$ P(Y_{tr}|V_{tr}) = \Sigma_{\pi_1'..\pi_{k'}'} P(Y_{tr}|\pi_1'..\pi_t')P(\pi_1'..\pi_{k'}'|V_{tr}, \overline{V_{tr}}) $$
$$
= \Sigma_{\pi_1',.\pi_{k'}'} P(Y_{tr}|\pi_1'..\pi_{k'}')
\Pi_{i}P(\pi_i'|S_i)
\Pi_{j} P(\pi_j|V_{tr}, \overline{V_{tr}})
$$


As we can see in the model, the utility of each trend depends on leaf policies $P(\pi_j|V_{tr}, \overline{V_{tr}})$, complex policies $P(\pi_i'|S_i)$ and conditional distribution of $Y_{tr}$ given complex policies. 
$P(\pi_i'|S_i)$ can be interpreted as the weight of complex policy $\pi_i'$.
Our goal is to find the structure and parameters of the model to maximize the probability of observed data i.e:
$$ P(Y|V) = \Pi_{tr \in ts} P(Y_{tr}|V_{tr}, \overline{V_{tr}})$$

Implementing this utility model requires addressing several probabilistic modeling tasks:
\begin{itemize}
    \item {\bf Latent variable learning:} determining the types and parameters of the simple policies
    \item {\bf Structure learning:} identifying the dependencies necessary to identify complex policies
    \item {\bf Parameter estimation:} finding the conditional probability distribution for the trend's inclusion in the summary, given the complex policies
    \item {\bf Inference:} determining whether a given trend will appear in a summary
\end{itemize}

In the following subsection we describe how we can learn this model. We define prerequisite learning sub-tasks similar to~\cite{babi} for learning the complete utility model. Once the underlying graphical model is learned, we can use it to assign utilities to the new trends.

\section{Inference and Learning for Utility Models}
In this section, we describe how we address several of the core learning tasks for our utility estimation model.
\subsection{Parameter estimation for leaf policies}
As mentioned in the previous section the utility of a trend is dependent on the combination of leaf policies. Therefore the first step in learning the utility model is to capture various leaf polices. In this paper, we focus on several predefined types of leaf policies and focus on learning their parameters.

Leaf policies are binary random variables that indicate whether features of a trend have a specific relation or not. Some of these policies also consider dependency of trend to other trends in their relation. 
The probability distribution of a leaf policy $\pi$ which its value for a trend $tr$ is determined by its feature vector $v_{tr}$ is characterized by an indicator function with parameters $A \in R^m$, $b \in R$.
$$ P(\pi_{A, b}| V): R^m \rightarrow [0, 1] $$
$$P(\pi_{A,b} | V = v_{tr}) \propto  I[A^Tv_{tr} + b \geq 0] $$ 
This value of a policy for $tr$ is true if $A^T v_{tr} + b\geq 0$ and false otherwise. In other words, it is parametrized by a linear separator in $R^m$ which gives high value to the points above the line. 
As an example let $\pi_j$ be the policy whose value is true for linear increasing trends and false for non-increasing trends. The $tr$ be a linear trend and $j^{th}$ element in its feature vector $v_{tr}$ denote its slope. $\pi_j$ can be represented with a one hot vector $A$, $b = 0$ where $A_j = 1$.\\
In another group of leaf policies, the value of policy for a trend $tr$ is based on its feature vector and its relation with other trends. We focus on the pairwise dependency among trends and later show that dependencies involving more trends can be captured by pairwise dependencies in this problem, though it might not be the most efficient solution. 
Let $\pi$ be a leaf policy such that its value for $tr$ is determined by feature vector $V_{tr}$ and feature vector of another trend $V_{tr'} \in \overline{V_{tr}}$. The distribution of $\pi$ is characterized by an indicator function with parameters $A, B \in R^m$, $c \in R$.
$$P(\pi_{A,B,c} |V, V'): R^m \times R^m \rightarrow [0, 1]$$
$$P(\pi_{A,B,c} | V, V' = v_{tr}, v_{tr'}) \propto I[A^Tv_{tr}+B^Tv_{tr'}+ c \geq 0] $$
$P(\pi_{A,B,c} | V_{tr}, V_{tr'})$ considers the dependency between $tr$ and $tr'$. This policy has a true value for $tr$ if $A^Tv_{tr}+B^Tv_{tr'}+ c \geq 0$. It can be interpreted as a linear separator in $R^{2m}$. Note that policies that only consider a single trend are a special case where $B = 0$,  but for simplicity we separated their representations. 
As an example let $\pi_k$ be the policy that has higher value for more recent trends and $k$ be the element of feature vector that indicates the time span of trends. $\pi_k$ can be represented with two one hot vectors $A$, $B$ where $A_k = 1$, $B_k = -1$ and $c=0$. 
Policies like $\pi_k$ are building blocks of more complex policies which consider dependencies among multiple trends e.g the policy $\pi'$ which prefers the most recent trend in a time series can be expressed as conjunction of $\pi_k$ with itself 
where the each of time it repeats, it contributes to the dependency of $tr$ with one of the trends in $\overline{V_{tr}}$. Therefore $\pi'$ assigns highest utility to the most recent trend.\\
As we mentioned above, we characterized leaf policies with linear separators. Therefore our goal is to find parameters of these linear separators $(A_i, B_i, c_i)$ such that the probability of observed data is maximized. We expect probabilistic linear classifiers such as logistic regression perfectly detect parameters of these separators. It is also possible to use Maximum likelihood to find the parameters of each leaf policy when the graphical structure of the model is known.
\subsection{Structure Learning}
In the proposed utility model, subsets of leaf policies are combined via different structures and create complex policies. Therefore each complex policy is dependent on a subset of the leaf policies. We assume complex policies can be modeled as the product of the constituent leaf policies. The structure of the dependencies among leaf and complex policies are unknown. The problem of finding conditional dependencies between variables, which represent edges in our graphical model, has been well-studied~\citep{drton2016structure}. Our utility model is a Bayesian network, hence we use available structure learning methods for tree structured Bayesian networks as the baselines. We use greedy search and Chow-Liu~\citep{1054142} for learning the utility model structure.
\subsection{Learning Utility}
The final step of learning the utility model is to find the probability distribution of $Y$ given complex policies. The table containing the conditional probability of $P(Y|\pi_1',..,\pi_{k'}')$ in the Bayesian network contains $2^m$ entries. Therefore its impractical to compute all values in the table. For learning this table, a possible approach is to learn a probabilistic classifier for $Y$ given $ [\pi_1',..,\pi_{k'}']$ as feature vector. One candidate for a probabilistic classifier is logistic regression. We can also use the naive Bayes assumption and train a naive Bayes classifier.\\
Once the structure and parameters of the model are learned we can infer utility of new trends using them. 
They utility of a trend $t$ with feature vector $v_{tr}$ can be computed as: 
$$P(y = 1|v_{tr}, \overline{v_{tr}}) = $$
$$\Sigma_{\pi_1', .., \pi_{k}'}P(y = 1 | \pi_1',.., \pi_{k'}')P(\pi_1'..., \pi_{k'}' | v_{tr}, \overline{v_{tr}})
$$
$$P(\pi_1', ..\pi_{k'}'|v_{tr}, \overline{v_{tr}}) = \Pi_{i=1}^{k'}P(\pi_1'|v_{tr}, \overline{v_{tr}})$$
Computing the probability of $Y$ for all possible values of complex policies which are the hidden variables is computationally expensive. Also the conditional probability table of $Y$ is unknown. We train a classifier for estimating the utility based on complex policies and use it instead of conditional probability table. A possible approach to compute utility in this scenario is to find values of $\pi_1', .., \pi_t'$ with the highest probability then compute probability of $Y = 1$
for that specific assignment of complex policies. 
$$q_1, q_2, .., q_{k'} = \argmax_{\pi_1', ..,\pi_{k'}'} P(\pi_1',.., \pi_{k'}'|v_{tr}, \overline{v_{tr}})$$
$$U(v_{tr}) = P(Y=1|v_{tr}, \overline{v_{tr}}) $$
$$ \approx P(Y=1|\pi_1', ..,\pi_j'=q_1,..,q_j)\Pi_j P(\pi_j'=q_1|v_{tr}, \overline{v_{tr}})
$$


\section{Experiment}
In this section, we describe the experiment we conducted to evaluate our proposed utility model. We use synthetic data in our experiments to check the applicability of our model since real data is noisy and using them adds more complexity to the problem at this stage. We created a synthetic dataset consisting of 2000 numerical time series. For generating each time series in our synthetic data, we randomly segmented the time span. Then we inserted a random linear trend for each time span by adding points in that linear trend with normal noise.
In each experiment scenario, we created a training set which contains feature vectors of the detected trends in training time series. Then we learned each part of the model separately and evaluated the overall performance of the system for baseline methods.
\subsection{Evaluation}
The real value of utility for each trend is not available. As mentioned in previous sections, utility of trends is used in selecting subset of trends to appear in the text and determining ranking among them. Therefore we evaluate our system using two different metrics which are Precision/Recall and Kendall Tau each of which  evaluate one aspect of utility. We also evaluate the subtasks separately.
\subsection{Experiment Scenarios}
In this section we describe our experiments. In each scenario, we assumed structure and parameter learning are done in isolation and we evaluated the inferred utilities of the learned utility model. 
\subsubsection{Learning Leaf-policy parameters}
In this experiment set, we assumed the model consists of a single leaf policy and repeated the experiment for example leaf policies from different leaf policy types introduced in \ref{policies}.
We tried to learn the parameters of that single leaf policy and infer the utility of trends. Since there is only one complex policy and one leaf policy in this case, no structure learning is required.  In this experiment, the baselines are probabilistic linear separators e.g logistic regression. We also evaluated the performance of non probabilistic classifiers in this case including decision tree, SVM. We describe the leaf policies used in this experiment in Table \ref{policies}. 

\begin{table}
\begin{tabular}{|c|c|}
\hline
    Policy Id & Policy Preference Description \\ \hline
     $\pi_1$ & increasing linear trend \\
     $\pi_2$ & slope of linear trend greater than a threshold \\
     $\pi_3$ & maximum trend\\
     $\pi_4$ & specific trend type  \\
     $\pi_5$ & more recent trends \\
     $\pi_6$ & greater spanning interval \\
     $\pi_7$ & more extreme jumps \\
     $\pi_8$ & different trend types \\
     \hline
\end{tabular}
\caption{Leaf policy ids and their description. The description column describes the condition when the value of leaf policy is 1. The first four leaf policies are defined for a single trend whereas the last four policies are defined over pair of trends. The second column describes the condition when the value of each policy is 1. e.g the value of $\pi_1$ is 1 when the given linear trend has positive slope. or the value of $\pi_5$ is 1 when the change of first given jump trend is greater than the second given jump trend. }
\label{policies}
\end{table}
Results of baselines in this experiment are shown in Table \ref{leaf_eval}. As we expected the probabilistic classifiers model the leaf policies perfectly. Therefore, almost perfect f1-score is achieved for all leaf policies. We use the trained logistic regression classifiers for leaf policies in this experiment for the second experiment.
\begin{table}
\centering
\resizebox{\linewidth}{!}{
\begin{tabular}{|c|c|c|c|c|c|c|}
    \hline
    \multirow{2}{*}{Policy} & \multirow{2}{*}{Metric} & Logistic & Naive & Decision & \multirow{2}{*}{SVM} & Weighted Logistic\\
    & & Regression & Bayes & Tree &  & Regression \\
    \hline
    \hline
    \multirow{2}{*}{$\pi_1$} &  Kendall & 0.99 & 0.92 & - & - & 0.99\\
    & F1-score & 1.0 & 0.89 & 1.0 & 1.0 & 1.0\\
    \hline
    \multirow{2}{*}{$\pi_2$} &  Kendall& 0.99 & 0.45  & - & - & 0.99\\
    & F1-score & 1.0 &  0.36 & 1.0 & &0.98\\
    \hline
    \multirow{2}{*}{$\pi_3$} &  Kendall & 1 & 1 & - & - & 1\\
    & F1-score & 1 & 1 & 1 & 1 & 1\\
    \hline
    \multirow{2}{*}{$\pi_4$} &  Kendall & 1 & 1 & - & - & 1\\
    & F1-score & 1 &  1&  1& 1 & 1\\
    \hline
    \multirow{2}{*}{$\pi_5$} &  Kendall & 0.99 & 0.96 & - & - & 0.99\\
    & F1-score & 1.0 & 0.91 & 0.99 & 1.0 & 1.0\\
    \hline
    \multirow{2}{*}{$\pi_6$} &  Kendall & 0.97 & 0.48 & - & - & 0.99 \\
    
    & F1-score & 1.0  & 0.37 & 1.0 & 1.0 & 0.98\\ \hline
    \multirow{2}{*}{$\pi_7$} &  Kendall & 0.99 & 0.95 & - & - & 0.99\\
    & F1-score & 0.98 & 0.85 & 0.99  & 1.0 & 0.97\\
    \hline
\end{tabular}
}
\caption{Evaluation of baselines in first experiment. }
\label{leaf_eval}
\end{table}

\subsubsection{Utility Estimation Experiment}    
In this experiment, we assumed to have multiple leaf policies. We also assumed to have multiple complex policies and their dependency structure to the leaf policies are known. Our goal was to estimate utility based on complex policies. We did not keep the conditional probability table of $Y$ given complex policies. Instead, we trained a probabilistic linear classifier to estimate the utility given complex policies. Note that, we used the trained classifiers of the previous experiment to find the value of complex policies. 
In the first two scenarios, the leaf policies have the same type, while in the rest scenarios leaf policies have different types. The complex policies along descriptions and dependent leaf policies are shown in table \ref{complex_policy}.
\begin{table}
\begin{tabular}{|c|c|c|}
\hline
    Policy & Description \\ \hline
     $p_1$ & linear \emph{and} increasing  \\ 
     $p_2$ & jump point \emph{and} downward \\
     $p_3$ & extreme point \emph{and} has high value \\ 
     $p_4$ & linear \emph{and} highest spanning interval \\ 
     $p_5$ & linear \emph{and} sharpest increase \\ 
     $p_6$ & jump point \emph{and} sharpest uptrend \\ 
     $p_7$ & jump point \emph{and} most recent \\ 
     $p_8$ & jump point \emph{and} unique \\ 
     $p_9$ & linear \emph{and} most recent \\ \hline 
\end{tabular}
    \caption{The complex policies and their descriptions. The second column describes the conditions of the input trend that causes the value of each complex policy to be 1. The descriptions around \emph{and} are leaf policies that each complex policy depends on.}
    \label{complex_policy}
\end{table}
The results of the second experiment are shown in table \ref{eval2}

\begin{table}[]
    \centering
    \begin{tabular}{|c|c|c|}
    \hline
        Complex Policies & F1-score & Kendall \\ \hline
        $p_1, p_2$ & 0.98 & 1 \\ \hline
        $p_1, p_2, p_3$ & 0.99 & 1 \\ \hline
        $p_1, p_4$ & 0.99 & 1 \\ \hline
        $p_5, p_6$ & 0.94 & 1 \\ \hline
        $p_3, p_5, p_7$ & 0.98 & 1 \\ \hline
        $p_3, p_5, p_8$ & 0.98 & 1 \\ \hline
        $p_4, p_5, p_9$ & 0.98 & 1 \\ \hline
        \end{tabular}
    \caption{Evaluation of logistic regression for estimating utility based on complex policies. The final policy is assumed to be proportional to disjunction of complex policies i.e the final policy in these experiments assigns same weight to complex policies.}
    \label{eval2}
\end{table}
As shown in table \ref{eval2}, when the complex policies and their correct values for a given trend are known estimating utilities by using a probabilistic linear classifier can achieve high score in many setting. However, we should note that the performance of utility estimator highly depends on the value of complex policies. 



\section{Previous Work}
Data-to-text systems have long been an area of active research. There have been various Data-to-text systems focusing on creating textual summary for different data. \cite{chartToText, Demir2012SummarizingIG} provide examples of data-to-text systems that focus on generating textual descriptions for graphical or chart data. 

A  data-to-text  system  represents  the  given  data or  knowledge  in a text  format  so  that  people  can  understand and interpret the information better. The workflow in a data-to-text system consists of modules that are responsible for analyzing the input data and extracting patterns and trends, detecting the relation between trends, selecting the content and generating the output~\citep{datatotext}. 

In this paper, we limit the domain of the system to numerical time series data. \cite{sripada-etal-2003-summarizing, sumtimeWeather} focus on creating description for time series data on different domains. Their systems depends on expert knowledge in the content selection phase. \citep{lloyd2014automatic} creates description for time series data by discovering statistical models in it and map them to natural language text for creating a good explanation of data. However the provided explanation consists of description of complicated statistical patterns such as "This component is a smooth function with a typical length scale of 8.1 months" which are not appealing for nontechnical reports and are not similar to human descriptions. 
Our approach follows the pattern of data-to-text systems: we create a data-to-text system to generate a  qualitative  summary  for  a  given  time  series. Our  goal is to provide enough information in the summary so that user can reason about a series based solely this summary, without requiring quantitative analysis.\\
Analyzing time series data and extracting trends and patterns from it which is the first component of our system have been studied in \citep{lloyd2014automatic, trendtemplate}, \citep{hwang2015a}. 
\\
The content selection component of data-to-text systems resembles the extractive document summarization problem. In extractive document summarization, the goal is to select a subset of the documents or patterns to represent the whole document. More precisely, these methods select top $k$ most important sentences in documents by greedy search or optimizing an objective function. \citep{Survey_Allahyari_2017}. Their objective function is usually a combination of a submodular and non submodular function that adjusts the redundancy and informativeness of the created summary. \citep{dasgupta-etal-2013-summarization, Lin:2011:CSF:2002472.2002537}. In this work, by estimating utility value for each trend we provide a means for selecting top $k$ trends in a time series. Estimating utility for trend also enables us to defind submodular objective functions for selecting a subset of trends.


\section{Conclusion}
In this paper, we introduced a model of simulating human like descriptions for time series data. Our initial work is focused on identifying substasks to learn such model. In our evaluation, we showed the result of baseline on substask and showed that although learning each subtask is straightforward, learning combination of them is a complex task. In our ongoing work, we are working to learn substasks simultaneously. 


\section{Acknowledgments}
This material is based upon work supported in part by United States Air Force and the Defense Advanced Research Projects Agency (DARPA) under Contract No. FA8650-17-C-7715 and award W911NF-18-1-0027, and is based on research sponsored in part by the National Science Foundation under grant OIA-1937153. 





\bibliographystyle{ACM-Reference-Format}
\footnotesize{
\bibliography{pegah-starai20}
}

\end{document}